\documentclass{article} 
\usepackage{listings, graphicx}
\usepackage{xcolor,soul}
\usepackage{hyperref}
\usepackage{float}

\usepackage{iclr2024_conference,times}

\usepackage{amsmath,amsfonts,bm}

\def\eqref#1{equation~\ref{#1}}

\def\1{\bm{1}}

\DeclareMathAlphabet{\mathsfit}{\encodingdefault}{\sfdefault}{m}{sl}
\SetMathAlphabet{\mathsfit}{bold}{\encodingdefault}{\sfdefault}{bx}{n}

\usepackage{hyperref}

\usepackage{url}

\title{NITRO: LLM Inference on Intel\textsuperscript{\textregistered{}} Laptop NPUs}

\author{Anthony Fei and Mohamed S. Abdelfattah \thanks{Thanks to Intel Labs for funding support.} \\
Cornell University \\
\texttt{\{ayf7,mohamed\}@cornell.edu} \\
}

\iclrfinalcopy 
\begin{document}

\maketitle

\begin{abstract}
Large Language Models (LLMs) have become essential tools in natural language processing, finding large usage in chatbots such as ChatGPT and Gemini, and are a central area of research. 
A particular area of interest includes designing hardware specialized for these AI applications, with one such example being the neural processing unit (NPU). 
In 2023, Intel released the Intel Core Ultra processor with codename Meteor Lake, featuring a CPU, GPU, and NPU system-on-chip. 
However, official software support for the NPU through Intel's OpenVINO framework is limited to static model inference.
The dynamic nature of autoregressive token generation in LLMs is therefore not supported out of the box.
To address this shortcoming, we present \textbf{NITRO} (\underline{\bf N}PU \underline{\bf I}nference for \underline{\bf Tr}ansformers \underline{\bf O}ptimization), a Python-based framework built on top of OpenVINO to support text and chat generation on NPUs. 
In this paper, we discuss in detail the key modifications made to the transformer architecture to enable inference, some performance benchmarks, and future steps towards improving the package. The code repository for NITRO can be found here: \url{https://github.com/abdelfattah-lab/nitro}.
\end{abstract}

\section{Introduction}


As Large Language Models (LLMs) continue to rise in popularity, so does the hardware they run on. GPUs are essential in running these models due to their high compute capabilities and parallelism for common vector and matrix operations present in transformers. NVIDIA GPUs are the most popular choice, with platforms such as CUDA and PyTorch making deep learning models convenient to develop and deploy \citep{DBLP:journals/corr/abs-1912-01703}, \citep{7476520}.

\subsection{Intel Meteor Lake chipset}

Recently, Intel has made efforts in specialized hardware design for machine learning tasks. In 2023, Intel released the Meteor Lake chipset, the first series of its Intel Core Ultra processors, featuring four distinct tiles that handle their own uses. The tiles include the compute, graphics, system-on-chip (SoC), and I/O tiles.

The system-on-chip (SoC) tile serves as the central coordinating unit, integrating essential processor components such as memory management, compute logic, and input/output (I/O) interfaces onto a single silicon tile. It employs a Network-on-Chip (NoC) and I/O Fabric to facilitate high-bandwidth and low-latency communication between the SoC and the other tiles.

The graphics tile is Intel's integrated GPU. The compute tile consists of most of the CPU cores (classified as P-cores and E-cores). The I/O tile handles the communication between devices. Figure 1 demonstrates the layout of these tiles, as well as the NoC and IO Fabric connecting these components. \citep{Bonshor24}

\begin{figure}[H]
    \centering
    \includegraphics[width=0.7\linewidth, page=1]{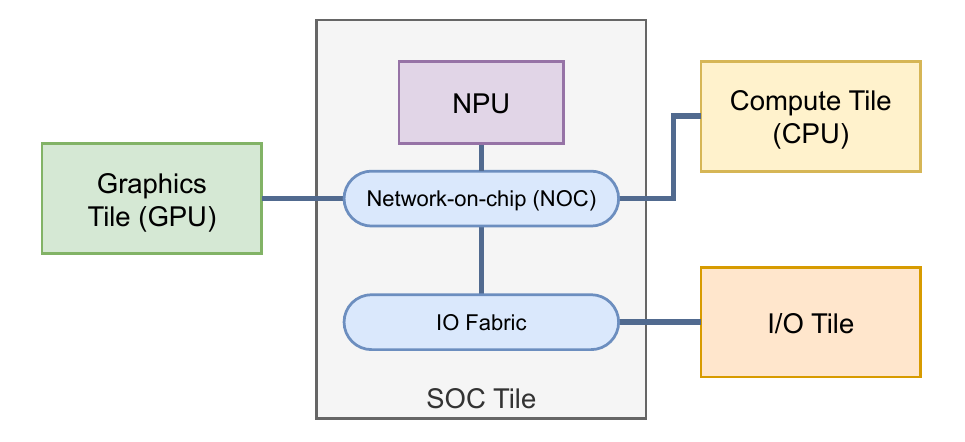}
    \caption{High level architecture of the Meteor Lake chipset.}
    \label{fig:enter-label}
\end{figure}

A particular block on the SoC is the neural processing unit (NPU), a device specialized on computations for AI tasks at low power. The Meteor Lake's NPU consists of two neural compute engines, each of which contains a Streaming Hybrid Architecture Vector Engine (SHAVE) DSP for parallel and general computing, and an inference pipeline handling the higher compute operations consisting of blocks designed for MAC units, activation functions, and data conversions \citep{Chester24}. With 4096 MACs operating at 1.4 GHZ, the NPU provides 11 TOPS. For comparison, the Meteor Lake CPU provides 5 TOPS and the GPU provides 18 TOPS \citep{Grunin23}.

The \href{https://docs.openvino.ai/2024/index.html}{OpenVINO toolkit} is a platform that supports deployment of machine learning models on Intel CPUs, GPUs, and NPUs. OpenVINO supports model deployment on the NPU, and has found use cases in computer vision applications such as YOLOv10. However, as it is a relatively new device, support is still limited: out-of-the-box, NPUs do not readily support LLM deployment.

\subsection{Objective}

Hosting LLMs on mobile computing machines is of great interest: being able to support these models locally means users will not have to heavily rely on servers for chatbot assistance, which often charge a price for tokens due to high compute power. It is also a challenging task, as top performing models generally will exceed the capabilities of mobile devices such as a laptop.

In this technical report, we will discuss the incompatibilities of the transformer architecture and the NPU as well as the current state of the OpenVINO Ecosystem. 
We will then present details about the implementation of our package NITRO, highlighting architectural changes to the transformer. Some benchmark results are shown in comparison to CPU and GPU compute, and further development and research areas are discussed. 

\section{The OpenVINO Ecosystem}
In this section, we discuss the current OpenVINO ecosystem, and the relevant frameworks and packages currently in production to serve machine learning models. For each framework, we underscore why LLM inference on NPUs is not supported out-of-the-box, or is difficult.

\subsection{OpenVINO}
OpenVINO is Intel's open-source toolkit that enables inference of machine learning models on Intel CPUs, GPUs, and NPUs. This toolkit has its own model representation called the OpenVINO intermediate representation (OpenVINO IR). There are two files that make up this model: \textbf{an .xml file}, which specifies the structure of the model, and \textbf{a .bin file}, which provides the weights of the model.

OpenVINO IR models resemble the structure of a direct acyclic graph, where nodes are operations (e.g. matrix multiplication (MatMul), softmax, ReLU) and edges are connections between operations. The key attributes of a node include:

\textbf{Inputs and Outputs.} Also referred to as ports, inputs define the properties of a tensor to be used as part of the operation. it defines the shape through the \texttt{<dim>} tag. Ports are differentiated by some index or optionally a name, and a precision for the data type. A dimension can also take on the special value \texttt{?}, which declares an unknown dimension. Tensor inputs with this value are known as dynamic tensors.

\textbf{Node Type.} In addition to standard operations such as MatMul, there are other special types of nodes. For instance, \texttt{Parameter} is a node that accepts inputs to the model, \texttt{Result} is a node that is the output to a model, and \texttt{ReadValue} and \texttt{Assign} are special operations that allow us to assign values to internal variables called states.

\textbf{Identifiers.} Every node is assign a unique number. Generally, the nodes are assigned in sequential order, with earlier operations having lower numbers and later operations having higher numbers. Nodes can also have a specific name, which makes accessing and modifying nodes a lot easier.

\begin{lstlisting}[language=XML, caption=Example node representation in OpenVINO.]
<layer id="1137" name="<...>" type="MatMul" version="opset1">
    <data transpose_a="false" transpose_b="true" />
    <input>
        <port id="0" precision="FP32">
            <dim>1</dim>
            <dim>1</dim>
            <dim>2048</dim>
        </port>
        <port id="1" precision="FP32">
            <dim>8192</dim>
            <dim>2048</dim>
        </port>
    </input>
    <output>
        <port id="2" precision="FP32" names="1425">
            <dim>1</dim>
            <dim>1</dim>
            <dim>8192</dim>
        </port>
</output>
</layer>
\end{lstlisting}

Listing 1 provides an example of such a node with id 1137, which performs matrix multiplication $AB^T$, where $A$ (on port 0) has shape $1 \times 1 \times 2048$ and $B$ has shape $8192 \times 2048$. It outputs a tensor of $1 \times 1 \times 8192$, with port name ``1425''. Because all dimensions are specific, this node is said to operate on static shaped tensors.

\begin{lstlisting}[language=XML, caption=Example edge representations in OpenVINO.]
<edge from-layer="1136" from-port="1" to-layer="1137" to-port="1" />
<edge from-layer="1137" from-port="2" to-layer="1138" to-port="1" />
<edge from-layer="1138" from-port="2" to-layer="1141" to-port="0" />
\end{lstlisting}

Similarly, the XML contains information about edges at the end of the file. In Listing 2, we see that the output tensor of the matrix multiplication operation defined in Listing 1 will be sent to the node with ID 1138.

The primary method of preparing a model in OpenVINO is through \textit{conversion} from an existing framework like PyTorch or TensorFlow to OpenVINO IR. Once a model has been prepared, it can be compiled to an Intel CPU, GPU, or NPU.

The NPU currently has a constraint in that it only supports static models: in its OpenVINO IR format, every node must have a defined shape.

This constraint is difficult to consolidate with the architecture of transformers models. Note that attention is defined by
\[ \sigma (\frac{QK^T}{\sqrt{d_k}}) V\]
where $\sigma$ is the softmax operation, $Q$ is the query projection of the input token(s), and $K$ and $V$ are the key and value projections of the \textit{past} token sequence. With each decode step of an LLM in text generation, new key and value projections are concatenated to $K$ and $V$. As a result, implementations of attention in OpenVINO generally operate on dynamic tensors.

In text generation, inference is also split into two parts: the pre-fill stage, where a prompt is inputted to generate the key and value projections in the KV-Cache, and the decode stage, where a new token is inputted to generate a probability distribution for the next token. The different input sizes propagate throughout the entire model, meaning that \textit{every} node in a transformer operates on dynamic shapes.

\subsection{Optimum-Intel}

\href{https://huggingface.co/docs/optimum/en/intel/index}{Optimum Intel} is an extension that provides conversion from Hugging Face transformers models into OpenVINO IR format. By default, these converted models are dynamic for reasons discussed from the past section. Out of the box, this extension enables LLMs on Intel CPUs and GPUs only.

\subsection{Intel NPU Acceleration Library}
The \href{https://github.com/intel/intel-npu-acceleration-library}{Intel NPU Acceleration Library} is an independent package designed to enable NPU inference for LLMs. Besides the \href{https://github.com/intel/intel-npu-acceleration-library/blob/main/src/bindings.cpp}{basic operation use} from OpenVINO as bindings, it mostly develops the infrastructure on its own. From benchmarks, NPU inference is rather slow compared to existing solutions, and is still in active development. 

\section{NITRO Framework}

\begin{figure}[b]
\begin{center}
\includegraphics[scale=0.7,page=2]{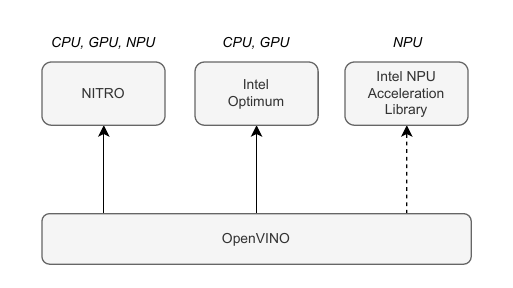}
\end{center}
\caption{OpenVINO Ecosystem.}
\end{figure}

\subsection{Objective}
The NITRO framework is motivated by three principles, from observations of the current software ecosystem and research landscape.

\textbf{Faithful to OpenVINO.} Because OpenVINO is in principle optimized for inference on Intel devices, we want the OpenVINO inference engine to do most of the work. Any additional code (which is run on CPU instead of NPU) will increase overhead.

\textbf{Flexibility.} The NPU is a product for Intel Laptops, and NITRO should be a product that can serve many use cases. Moreover, with a lot of ideas in the research field, we want to develop good software that can support novel techniques.

\textbf{Easy to maintain.} With many frameworks, it can be hard to understand what is going on. For instance, many Hugging Face models have so many parameters it can be difficult to distinguish which parameters are being utilized. Optimum Intel and Intel NPU Acceleration Library also are quite involved. A package that is easily maintainable enables more contributions and furthering the goals of our package.

These principles are interdependent: with high flexibility, it is easier to maintain. Staying faithful to OpenVINO and keeping the framework as light as possible reduces the amount of maintenance required.

\subsection{Overview}

NITRO obtains statically shaped transformer models through the conversion of custom PyTorch models into OpenVINO IR. These PyTorch models were modified from existing implementations taken from Meta's Llama \href{https://github.com/meta-llama/llama-models/blob/main/models/llama3/reference_impl/model.py}{code repository} and Hugging Face. A custom converter module is utilized to convert the PyTorch model into OpenVINO IR in a specific way due to obstacles with a direct conversion. Finally, NITRO contains its own inference pipeline for streamlined text generation, providing a user interface similar to Hugging Face. These three components---rewritten models, conversion, and pipeline---will be discussed in the following sections.

\begin{figure}[H]
\begin{center}
\includegraphics[scale=0.4,page=5]{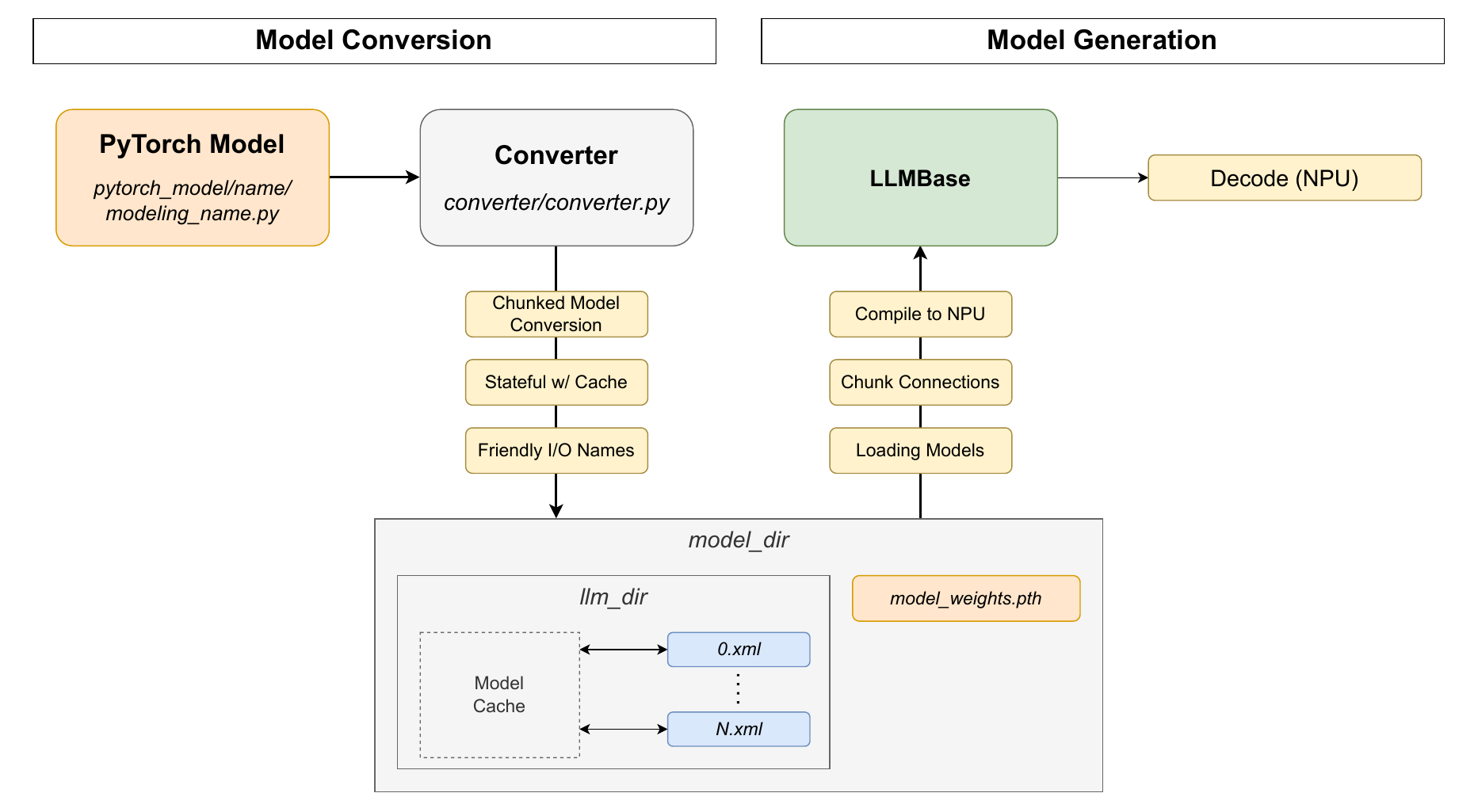}
\end{center}
\caption{The NITRO Framework.}
\end{figure}

\subsection{Rewritten PyTorch Models}

As mentioned in 2.5, existing implementations of LLMs in OpenVINO are dynamic. With PyTorch models, this is true as well: in Hugging Face, new key and value projections are concatenated to the KV-Cache, changing the shape of intermediate tensors with each iteration. 

\begin{lstlisting}[language=Python, caption=Snippet of the DynamicCache implementation in Hugging Face.]
 def update(
        self,
        key_states: torch.Tensor,
        value_states: torch.Tensor,
        layer_idx: int,
        cache_kwargs: Optional[Dict[str, Any]] = None,
    ) -> Tuple[torch.Tensor, torch.Tensor]:
    ...
    if len(self.key_cache) <= layer_idx:
        # There may be skipped layers, fill them with empty lists
        for _ in range(len(self.key_cache), layer_idx):
            self.key_cache.append([])
            self.value_cache.append([])
        self.key_cache.append(key_states)
        self.value_cache.append(value_states)
\end{lstlisting}

As a result, we re-implement the family of Llama models entirely, which are located at \href{https://github.com/abdelfattah-lab/nitro/tree/main/nitro/pytorch_model}{\texttt{nitro/nitro/pytorch\_model}} in the repository. Compared to Hugging Face models, the following changes were made:

\textbf{All inputs are tensors.} Hugging Face transformer models make excessive use of non-traceable inputs. This makes a direct conversion from Hugging Face to OpenVINO difficult: the conversion module in native OpenVINO utilizes TorchSript tracing, where primitive types and excessive use of None often results in unsuccessful conversions. By setting all inputs to tensors, this enables an easy conversion to OpenVINO IR.

\textbf{Extended and fixed-size KV-Caches.} To address the dynamic nature of the KV-Cache, we extend the KV-Cache to contain additional, unused slices (initialized with zeros) up to the maximum sequence length. The attention mechanism is then computed on the \textit{entire} $K$ and $V$ matrices, including the unused slices.

More concretely, let $K$ and $V$ be the original key and value tensors of the past tokens, and $m$ be the maximum number of tokens in text generation: with $n$ tokens in the present context, then for simplicity represent $K$ and $V$ as $n \times d$ matrices. Define $K'$ and $V'$ as $m \times d$ matrices, where the first $n$ rows are the same rows as $K$ and $V$ and the last $m-n$ rows are zeros, and $d$ is some hidden dimension size. Let $Q$ be a $1 \times d$ matrix representing the new token embedding.

Then $QK'^T$ is a $1 \times m$ matrix, where the first $n$ columns are the same as $QK^T$, and the last $m-n$ are filled with zeros. To correctly compute softmax, we need to mask the last $m-n$ entries with $-\inf$: let $M$ be the the attention mask, with dimension $1 \times m$ with the first $n$ entries being zero and the last $m$ entries being $-\inf$. Then the first $n$ entries in $\sigma(\frac{QK'^T }{\sqrt{d}}+ M)$ contains the same values as $\sigma(\frac{QK^T }{\sqrt{d}})$, and the last $n-m$ entries are filled with zeros. The final matrix multiplication with $V'$ then gives the same output as the original computation.

\textbf{KV-Caches as Input.} While Hugging Face models provide cache utilities, representing caches as a class, at its core it is simply a tensor value of the past sequence key and value projections.

This alone will result in incorrect computations due to the softmax function. To address this, we also extend our attention mask to include $-\inf$ along slices that are unused: when the softmax function is then utilized, these slices are set to 0 after exponentiation and will not be considered in the computation.

\textbf{Rotary Embeddings as Input.} The rotary embeddings were moved as an input into the function due to the dynamic nature of slicing the precomputed embeddings with each position value. In particular, we compute the function in the Llama repository, \texttt{precompute\_freqs\_cis} outside the function and then feed the corresponding slices of each position into the model. Further testing can be done as to effectively re-incorporate this into the OpenVINO IR.

Figure 3 outlines the design of the decoder layers.

\begin{figure}[t]
\begin{center}
\includegraphics[scale=0.6,page=3]{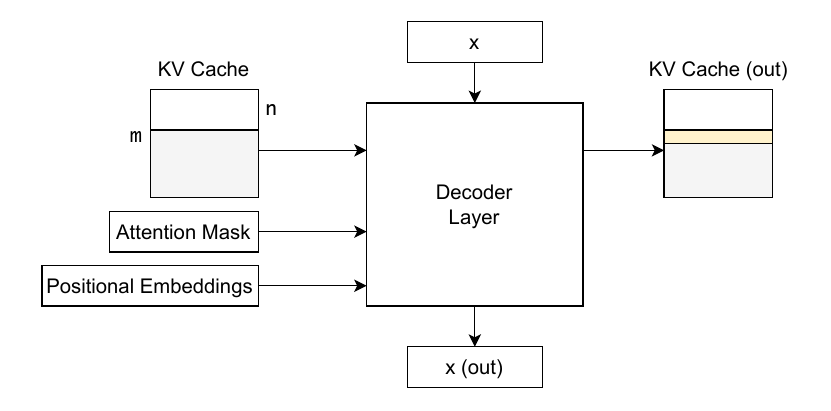}
\end{center}
\caption{Diagram of the decoder layer design. In addition to the token embedding, the KV-Cache, mask, positional embeddings are inputs to the layer. The output is the next token embedding as well as the present KV-Cache, where a new slice (in yellow) is updated.}
\end{figure}

\begin{figure}[t]
\begin{center}
\includegraphics[scale=0.9,page=4]{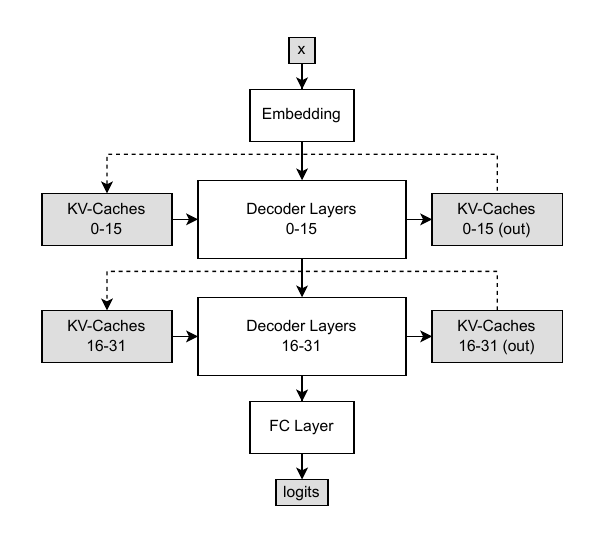}
\end{center}
\caption{Chunking structure of the model: white represents the OpenVINO IR, and gray represents the input/outputs (attention mask and positional embeddings not represented). Because the KV-Cache output is immediately fed back into the model, we can think of the cache values as an internal variable respective to each decoder layer.}
\end{figure}

\subsection{Model Conversion}

With the modified transformer architecture, we can now utilize OpenVINO's conversion method to convert the model into OpenVINO IR form, enabling the NPU. However, from tests we found that the compilation of NPU quickly exceeded our memory footprint. Attempts to compile the 8 billion parameter Llama 3 exceeded 96 GB.

\textbf{Chunking.} A key observation is that decoder-only models are very much sequential: much of the decoder-only transformer consists of a sequence of decoder layers: attention followed by feed forward. As a result, it is very easy to split the model. This motivates ``chunking" the PyTorch model by converting blocks of the model at a time, and clearing memory after each conversion. This allows the entire model to be converted without using too much memory.

A simple chunking pattern we utilize includes converting the embedding and final fully-connected layer on their own, and splitting the decoder layers into equal parts. For instance, Llama3-8B contains 32 decoder layers. A chunking pattern includes the embedding layer, the first 16 decoder blocks, the last 16 decoder blocks, and the final fully connected layer, giving a total of four OpenVINO IR models.

\textbf{Naming Convention.} As previously covered in section 2.1, in addition to integer identifiers nodes also can be assigned names which they can be accessed by. Having descriptive names is particularly important for Parameter and Result nodes, where the lack of descriptive names can be especially problematic with the KV-Cache inputs and outputs, where the tensors all have the same shape and the order of input/output nodes are not well-defined.

The converter module renames all parameter and result nodes of each chunk with the following convention: \texttt{token} is the input token into the embedding layer. All hidden states values, which include the output of the embedding layer, inputs and outputs to the decoder layers, and the input to the final FC layer, are named \texttt{x}. The final output is named \texttt{logits}. The KV-Cache inputs are named \texttt{cache\_v\_\textless x\textgreater } and \texttt{cache\_k\_\textless x\textgreater}, where \texttt{x} is the decoder block number (zero-indexed). \texttt{freqs\_cis} are the rotary embeddings.

\textbf{Stateful KV-Caches.} In section 3.3, we deliberately set the KV-Cache tensors as an input/output. This is our motivation: because the outputs of the KV-Caches are immediately fed back into the inputs, these values can be thought of as internal states to the model. OpenVINO provides an API that allows Parameter/Result pairs to be converted into ReadValue/Assign pairs. After chunking, each Model has its the KV-Caches converted into states, initialized as zero.

\subsection{Model Directory Structure}

Models are converted and exported to a directory with a standardized structure. If \texttt{model\_dir} is the specified model directory, then \texttt{model\_dir/llm\_dir} contains the OpenVINO IR models. Outside this directory, \texttt{model\_weights.pth} contains the PyTorch weights to the model, which are imported from Hugging Face and re-saved. Finally, \texttt{model\_dir/llm\_dir/cache} contains the cache for faster NPU compilation. This structure is illustrated in figure 3.

\subsection{Model Inference}

Finally, we set up a standard pipeline to generate text in Python, abstracting away the setup for text generation. 

\textbf{LLMBase.} The LLMBase re-connects and abstracts all the chunked OpenVINO IR models together. After initialization, the \texttt{\_\_call\_\_} function is used to run inference on the model with provided input tensors, which will return the results.

\textbf{LLMPipeline.} The LLMPipeline is in charge of the decoding process. In addition, the rotary embeddings and attention masks are managed at this level. It wraps an LLMBase object and can be used to generate text or act as a chatbot. There are three main functions in consideration: \textit{from\_pretrained,} which loads the weights from Hugging Face into the PyTorch model and converts the model into a specified directory, \textit{generate,} which takes in a prompt and generates tokens similar to Hugging Face, and \textit{chat\_generate,} which sets up a chat bot with a specified instruction model.

\section{Algorithms and Benchmarks}

To test the performance of different models, we utilize a laptop with a Meteor Lake Intel Core Ultra processor. The computer has 96 GB RAM, and we set up Ubuntu 22.04 LTS with kernel 6.9.3. We utilize the \href{https://github.com/intel/linux-npu-driver/releases}{Linux NPU Driver} version 1.10.0, and OpenVINO 2024.4.0. We consider different model configurations and benchmark the speed of NPU inference, and compare with the CPU and GPU.

It is worth noting that when benchmarking the GPU, the outputs are incorrect; text generation resulted in tokens that were not coherent. This started happening roughly around when OpenVINO 2024.4 was released. Different Intel Graphics Compute Runtime versions were tested, including reverting back to an older version, but we were unable to restore the correct computations.

\subsection{Baseline Comparison}

We compare different sized models, including Llama3.2-1B, Llama3.2-3B, and Llama3-8B. We compile the each model onto the CPU and GPU, and measure the average inference time per token. Note that for the CPU and GPU we utilize the same static model setup. The chunking configuration is the same as outlined in section 3.4, with one chunk of all 16 decoder layers in Llama3.2-1B, two chunks of 14 layers each in Llama3.2-3B, and two chunks of 16 layers each in Llama3-8B.

\begin{table*}[t]
\caption{Average decode inference times (ms/token) on different devices.}
\label{sample-table}
\begin{center}
\begin{tabular}{cccc}
\multicolumn{1}{c}{\bf Device} &\multicolumn{1}{c}{\bf Llama3.2-1B} &\multicolumn{1}{c}{\bf Llama3.2-3B} &\multicolumn{1}{c}{\bf Llama3-8B} \\
\hline
CPU    & 103.5 & 267.2 & 635.8 \\
GPU    & 60.4 & 132.4 & 273.6 \\
NPU    & 80.3 & 156.1 & 339.2 \\
\end{tabular}
\end{center}
\end{table*}


\subsection{Max Sequence Length}

Because we are running matrix multiplication across the max sequence length for all iterations, the number of computations is significantly greater compared to a dynamic model. On Llama-3-8B, we configure the maximum input size to different powers of 2, from 128 to 2048, measuring the average inference speed on CPU, GPU, and NPU.

\begin{figure}[t]
\begin{center}
\includegraphics[scale=0.7,page=6]{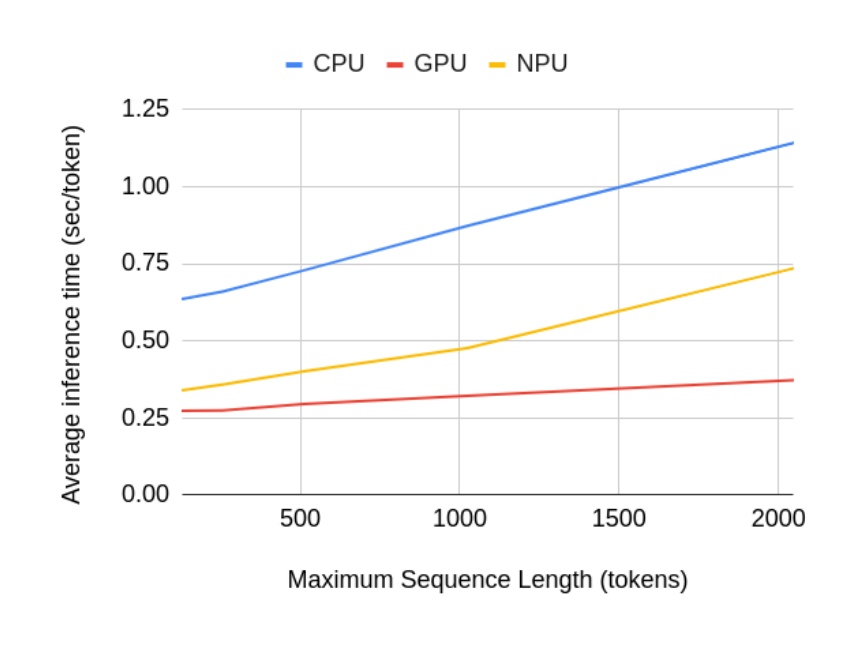}
\end{center}
\caption{Decode inference times, in seconds per token, on Llama3-8B, with different weight compression types.}
\end{figure}

From Figure 6, we observe that the NPU scales about the same as the CPU. The GPU exhibits superior performance across all sequence lengths. From 128 to 2048, a $16 \times$ increase in max sequence length, the GPU's inference increased by 0.0994 seconds/token, while the CPU and NPU's inference increased by 0.506 and 0.396 seconds/token, respectively.


\subsection{Quantization}

\href{https://github.com/openvinotoolkit/nncf}{NNCF} is supported for OpenVINO models. We test four levels of weight compression on Llama3-8B: INT8 ASYM, INT8 SYM, INT4 ASYM, and INT4 SYM, and observe the speedup across CPU and GPU. From Figure 7, while the CPU and GPU both improve their average inference speed, with times between 0.10 - 0.20 seconds/token, the NPU did not exhibit any improvement. Inference on the NPU exhibited no difference with INT8 ASYM and \textit{slowed down} with INT4 ASYM, meanwhile with both symmetric configurations, the compilation of the model failed. Listing 4 outputs the errors thrown for INT8\_SYM, with Linux NPU Driver 1.10.0, on OpenVINO 2024.4.0:

\begin{lstlisting}[caption=Error output on INT8\_SYM.]
Exception from src/inference/src/dev/plugin.cpp:53:
Exception from src/plugins/intel_npu/src/plugin/src/plugin.cpp:697:
Exception from src/plugins/intel_npu/src/plugin/src/compiled_model.cpp:63:
Exception from src/plugins/intel_npu/src/compiler/src/zero_compiler_in_driver.cpp:885:
L0 pfnCreate2 result: ZE_RESULT_ERROR_UNKNOWN, code 0x7ffffffe
\end{lstlisting}

Listing 5 outputs the errors thrown for INT4\_SYM:

\begin{lstlisting}[caption=Error output on INT4\_SYM.]
terminate called recursively
terminate called recursively
terminate called after throwing an instance of 'vpux::Exception'
terminate called after throwing an instance of 'vpux::Exception'
terminate called recursively
terminate called recursively
Aborted (core dumped)
\end{lstlisting}

\subsection{Comparisons to Other NPU Acceleration Libraries}

We also test the same setup using Intel NPU Acceleration Library. On both quantized INT8 and FP32/16 versions of Llama3-8B, we observe $\sim$3500 ms/token, so our platform provides over $10 \times$ speedup.

We also test the \href{https://docs.openvino.ai/2024/get-started/install-openvino/install-openvino-genai.html}{OpenVINO GenAI extension} for LLM deployment on NPUs. However, with Optimum 1.20 and OpenVINO 2024.5.0 we were unable to run inference: at runtime, it simply outputted a segmentation fault.


\begin{figure*}[t]
\begin{center}
\includegraphics[scale=0.7,page=7]{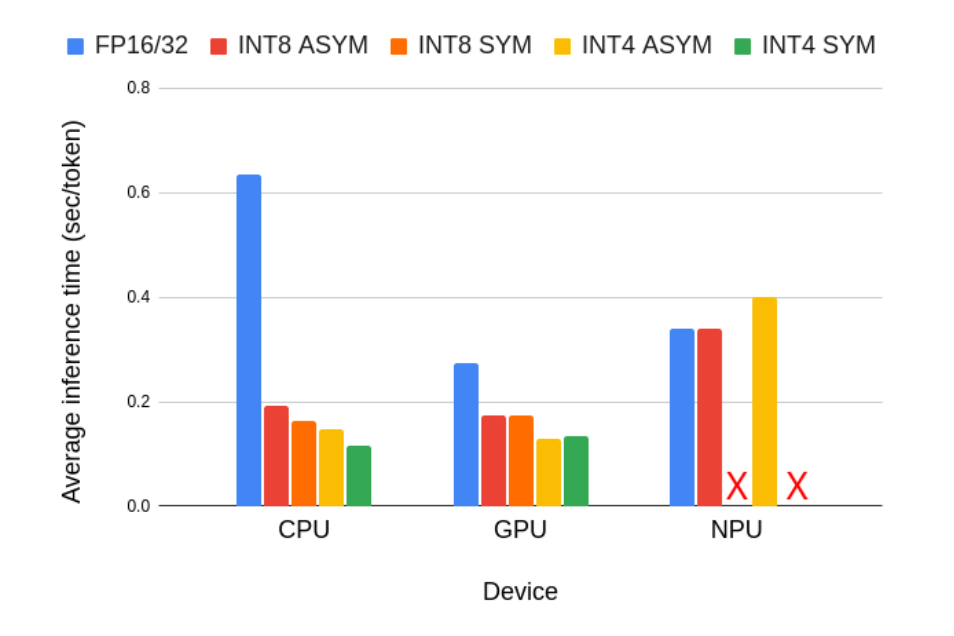}
\end{center}
\caption{Decode inference times, in seconds per token, on Llama3-8B, with different weight compression types. X's are placed on spots where an error occurred.}
\end{figure*}


\section{Areas of Improvement}

While NITRO enables the transformer architecture to be compiled and run on the NPU, there are improvements that certainly can be made to the framework.

\textbf{Rotary Embeddings in OpenVINO.} As previously mentioned, there are some features of the transformer architecture that are managed outside of OpenVINO, namely the rotary embeddings and attention mask. Operations that are done outside the model are through PyTorch or NumPy via CPU, which may increase overhead of the generation process. As mentioned, we want to remain as ``faithful'' to OpenVINO as possible, and the more we can offload generation processes to the OpenVINO models themselves, the better.

\textbf{LLM Optimizations.} A goal for NITRO is to implement novel techniques to provide additional speedup. A promising path is \textit{speculative decoding}: assuming that the number of tokens generated by the draft model is constant with each loop, we can fix the target model to accept a fixed number of tokens \citep{leviathan2023fastinferencetransformersspeculative}. This opens the opportunity of exploring heterogeneous computing, with compilation of the draft and token models on different devices. Pruning-based optimizations such as DejaVu may not be as viable, because it is not possible to remove specific operations in the model at runtime: the model is always compiled beforehand \citep{pmlr-v202-liu23am}.

\section{Conclusion and Further Research}

From running inference of transformers on the NPU, we observe some key limitations. Requiring a statically shaped model wastes a lot of compute because the unused slices from the key and value projects of the past sequence, and so each iteration of the decode stage runs the worst case every time. When keeping everything constant about our models, we find that the NPU is faster than the CPU by up to $1.8\times$ on intermediate sized models, such as Llama-3-8B, and the GPU remains faster.

However, there are NPUs exhibits restrictions compared to the CPU and GPU. The NNCF framework, while supposedly supporting OpenVINO models with weight compression and quantization, did not provide an improvement to the NPU specifically. This makes the CPU and GPU much faster alternatives, simply due to the availability of weight compression. More research needs to be done as to why this is the case, whether it is from the hardware itself or a bug with the software packages. Finally, the NPU's scaling being comparable to the CPU may suggest that the NPU may simply not be large enough to handle larger models, such as Llama-3-70B, compared to a traditional GPU.

The Lunar Lake processor chip is the successor to Meteor Lake, featuring a more capable NPU model with six compute engines compared to Meteor Lake's two. Testing NITRO on the Lunar Lake is another next step, and can give better context as to the poor scaling of the NPU inference.

Across all model sizes, The GPU generally demonstrates faster performance in terms of raw compute speed. However, the NPU is designed as a lightweight, energy-efficient alternative that provides a tradeoff between performance and power consumption \citep{Chester24}. While this paper focuses on implementing transformers optimized for NPU inference, one potential advantage of NPUs is their energy efficiency: they are reported to consume significantly less power compared to GPUs, making them a compelling choice for scenarios where energy savings are prioritized over peak performance.

\bibliography{iclr2024_conference}
\bibliographystyle{iclr2024_conference}

\end{document}